\def\CaptureBranding{1}
\documentclass{article} 
\usepackage{iclr2027_conference,times}

\usepackage{amsmath,amsfonts,bm}

\def\eqref#1{equation~\ref{#1}}

\def\1{\bm{1}}

\DeclareMathAlphabet{\mathsfit}{\encodingdefault}{\sfdefault}{m}{sl}
\SetMathAlphabet{\mathsfit}{bold}{\encodingdefault}{\sfdefault}{bx}{n}

\renewcommand{\eqref}[1]{(\ref{#1})}

\usepackage{hyperref}
\usepackage{url}
\usepackage{graphicx}
\usepackage{booktabs}
\usepackage{amsmath}
\usepackage{amssymb}
\usepackage{xspace}
\usepackage{multirow}
\usepackage{wrapfig}
\usepackage{float}
\usepackage[table]{xcolor}
\usepackage{needspace}
\usepackage{placeins}

\makeatletter
\def\section{\@startsection{section}{1}{\z@}{-1.4ex plus -0.4ex minus -.2ex}%
  {1.0ex plus 0.2ex minus 0.1ex}{\large\sc\raggedright}}
\def\subsection{\@startsection{subsection}{2}{\z@}{-1.2ex plus -0.4ex minus -.2ex}%
  {0.5ex plus .1ex}{\normalsize\sc\raggedright}}
\def\paragraph{\@startsection{paragraph}{4}{\z@}{1.0ex plus 0.3ex minus .2ex}%
  {-1em}{\normalsize\bf}}
\makeatother

\definecolor{oursblue}{HTML}{DCE9F7}
\newcommand{\ours}{\rowcolor{oursblue}}

\newcommand{\skytopia}{\textsc{Skytopia}\xspace}

\title{Skytopia: Monocular Drone Navigation with\\
Action-Conditioned Latent World Models}

\ifdefined\CaptureBranding
\iclrfinalcopy
\makeatletter
\newcommand{\fnmarkwidth}{\def\@makefnmark{\textsuperscript{\@thefnmark}}}
\makeatother
\author{\fnmarkwidth Yuhang Zhang$^{1}$\thanks{Equal contribution.},
Rangya Zhang$^{1}$\footnotemark[1],
Yujing Shang$^{1}$,
Zhuoyuan Yu$^{1}$,
Weiying Wang$^{2}$, \\
\bf\fnmarkwidth Steven Yang$^{2}$,
Qingsong Yan$^{3}$,
Chao Yan$^{4}$, and
Mir Feroskhan$^{1}$\thanks{Corresponding author.} \\
\small $^{1}$Nanyang Technological University, Singapore \quad $^{2}$Autel US \quad $^{3}$XGRIDS \quad $^{4}$Independent Researcher}
\else
\author{Anonymous Authors \\
Anonymous Institution \\
\texttt{anonymous@example.com}}
\fi

\begin{document}

\maketitle
\ifdefined\CaptureBranding
\lhead{}
\renewcommand{\headrulewidth}{0pt}
\vspace{-0.27in}
{\centering\small\textbf{Project page:}
 {\hypersetup{pdfborder={0 0 0}}\href{https://eugshang.github.io/Skytopia/}{https://eugshang.github.io/Skytopia/}}\par}
\vspace{0.07in}
\fi

\begin{abstract}
Monocular drone navigation requires reaching a goal in an unseen environment from a single
forward-facing camera, which offers few cues for depth and scale. World models address this by
modelling how observations evolve under actions, but they are built to be executed: the prediction
is produced at deployment and fed back into action generation at every control step. We argue that
what a policy needs from a world model is not the prediction but the representation required to
produce it: in flight the executed action explains almost all of the change between observations, so
prediction reduces to reprojecting a static scene under a known displacement. We therefore introduce
\skytopia, a policy built on an action-conditioned latent world model, and the 3D Gaussian Splatting
platform on which it is trained. A forward objective predicts the representation of the next
observation from the intended motion, and an inverse objective recovers that motion from the
predicted transition. Because the prediction never reaches action generation, the predictor is
discarded and one policy serves point-goal, image-goal, and goal-free navigation.
Simulation experiments show that \skytopia outperforms every baseline under
all three specifications, attaining 57.8\%, 66.0\%, and 49.0\% success rate, while discarding the
predictor removes 59.4\% of the inference cost. The same policy is subsequently deployed on a physical drone
without fine-tuning and reaches goals in indoor, open outdoor, and woodland environments.
\end{abstract}

\begin{figure}[H]
\centering
\includegraphics[width=0.91\textwidth]{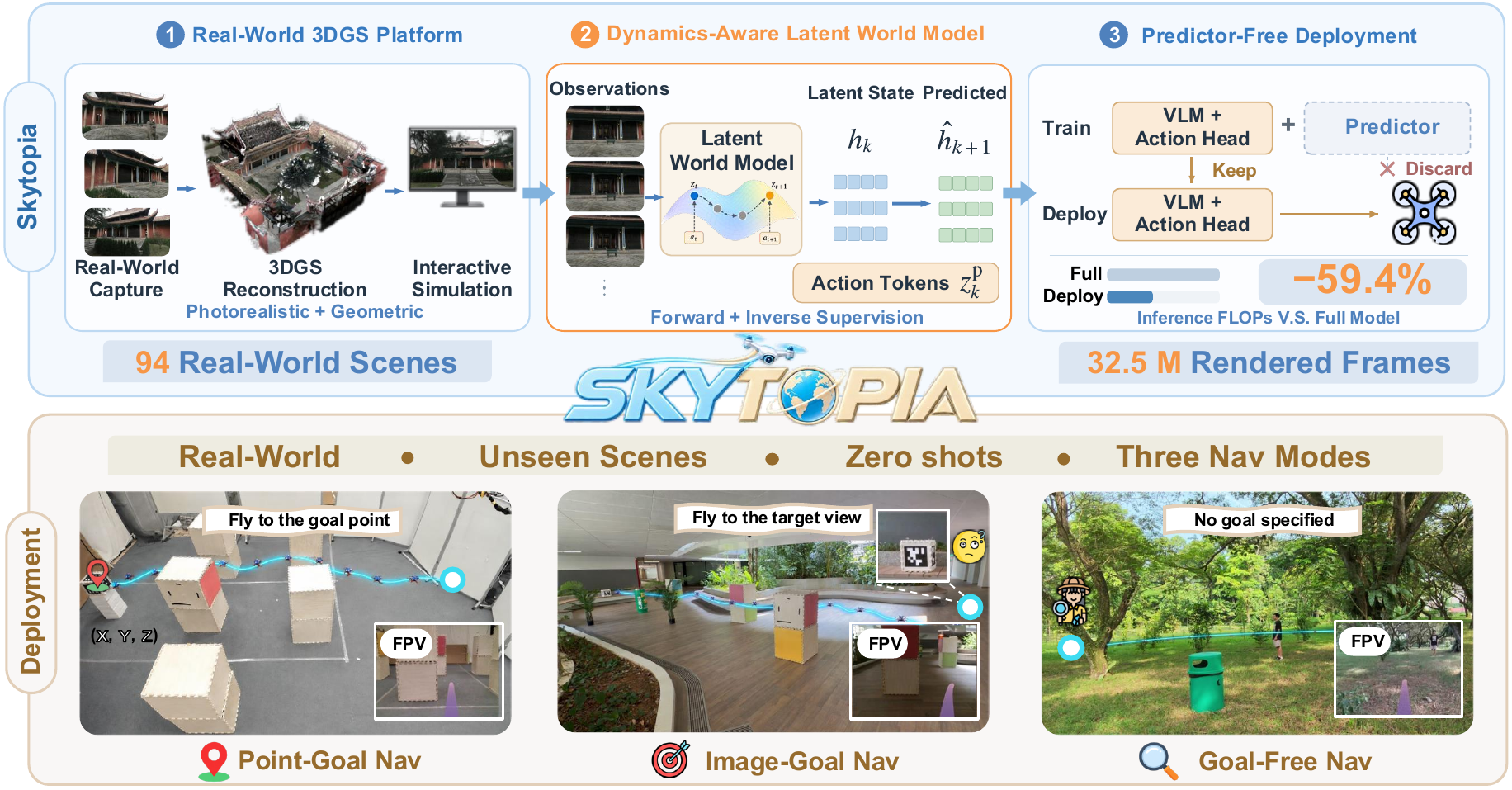}
\caption{\textbf{The \skytopia pipeline.} \emph{Top:} (1) real captures are reconstructed into a 3D
Gaussian Splatting platform of 94 scenes with rigid-body physics; (2) an action-conditioned latent
world model is trained on that platform under a forward and an inverse objective; (3) the predictor
is discarded once training is complete, removing 59.4\% of the inference cost. \emph{Bottom:} the resulting policy is evaluated zero-shot on a physical drone under the three goal specifications.}
\label{fig:cover}
\end{figure}

\section{Introduction}
\label{sec:intro}

Navigation is a foundational capability for embodied agents. It requires an agent to reach a
specified objective in an environment while producing smooth and collision-free
trajectories from visual observation. Aerial vehicles face the hardest form of this problem. Their
workspaces are open and weakly textured, and a single forward-facing camera provides few reliable
cues for depth and scale. Perception is therefore the dominant bottleneck. Human pilots nevertheless
fly first-person-view drones through cluttered scenes at high speed from exactly such a video stream
\citep{loquercio2021wild,kaufmann2023champion}. This raises the question of whether the same
competence can be conferred upon a policy that observes only what the pilot observes.

The most recent answer comes from world models \citep{hafner2025mastering,ding2025understanding,hou2026world}, which supervise the action together with how the
observation evolves once that action is executed. In flight the executed action accounts for most of
the change between consecutive observations, since the viewpoint moves far faster than the scene
itself changes. Prediction therefore reduces largely to reprojecting the scene under a known
displacement. This reprojection is governed by depth, which is precisely the quantity that a single
camera fails to provide. Learning to predict under the action thus amounts to learning the geometry
on which monocular flight depends. Ground navigation imposes a weaker form of this requirement. A
ground robot moves through structured and richly textured surroundings, where appearance alone
carries much of the information that navigation requires. Its camera also stays level, whereas every
acceleration of a multirotor tilts the camera and thereby rotates the viewpoint and blurs the image.
Finally, a ground robot can remain stationary while a prediction is computed. Existing navigation
world models are built for this setting and take two forms.
Explicit ones
\citep{bar2025nwm,luo2025grounding,chen2026imaginav} generate the frames that the agent would
perceive and learn the dynamics in pixel space. Latent ones
\citep{yao2025navmorph,chahe2026policy,zhang2026rae} predict a compact representation of the future
and plan over the imagined rollout at a far lower cost. In this setting the prediction is never
forced to depend on the action, and both forms rely on two design choices that flight cannot
afford.

\textbf{(i) The prediction is only weakly conditioned on the action.} Consecutive observations are
highly correlated. A predictor therefore lowers its loss by extrapolating the past and consults the
action only for a small residual. The resulting futures are visually coherent yet insensitive to the
executed action \citep{bar2025nwm,luo2025grounding,chen2026imaginav}. On the ground this is
tolerable, because appearance alone is often sufficient for navigation. In flight the relation between the action and the resulting image motion is the only reliable
source of depth. A predictor that ignores the action encodes how the scene looks over time rather
than how the action transforms it. The representation then lacks the depth and scale that obstacle
avoidance requires. \textbf{(ii) The predictor is retained at deployment.} Both forms produce the
prediction online and feed it back into action generation. A full world model hence executes at
every control step \citep{bar2025nwm,yao2025navmorph,chahe2026policy}. A ground robot can wait for this
computation. A drone cannot pause, and the inference latency is converted
directly into distance flown without a new command.

\begin{figure}[t]
\centering
\includegraphics[width=0.85\textwidth]{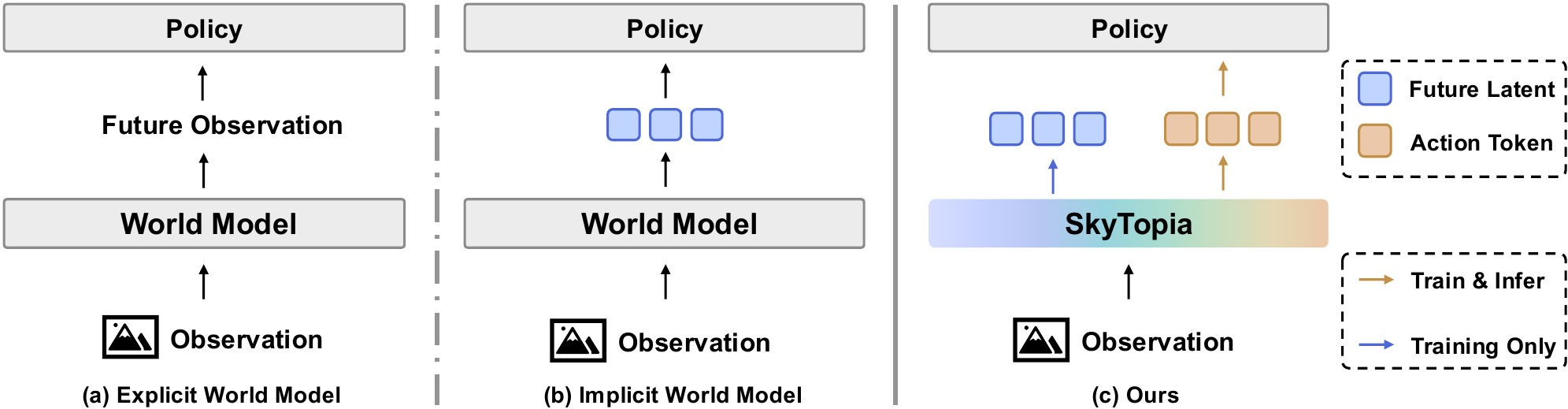}
\caption{\textbf{Where the prediction is used.} An explicit world model generates the next
observation and a latent one predicts its representation. In both cases the prediction is passed to
the policy at every control step. \skytopia produces the prediction during training only. It never
enters action generation, and the predictor is discarded at deployment.}
\label{fig:difference}
\end{figure}

In this paper, we propose \skytopia, shown in Fig.~\ref{fig:cover}, a unified framework for monocular drone navigation that consists of a simulation platform and an action-conditioned latent world model policy trained on it. \skytopia comprises three
components. \textbf{(1) A platform that supplies aerial data.} Demonstrations from the aerial
viewpoint are scarce, and a physical drone cannot provide them at comparable cost. The \skytopia
platform reconstructs real scenes as 3D Gaussian Splatting (3DGS) \citep{kerbl20233dgs} and couples
them with rigid-body physics, which makes photorealistic rendering and physical interaction
available in the same environment. From it we collect 32.5 million frames across 94 indoor and
outdoor scenes. \textbf{(2) A world model driven by the intended motion.} The backbone is presented
with a group of learned queries whose outputs are action tokens that summarise the intended motion.
A predictor estimates the representation of the future observation from those tokens, and an
inverse decoder recovers the executed motion from the resulting transition. The inverse objective
closes the shortcut of extrapolating the past. Because the predicted transition must retain the
command, the backbone cannot satisfy both objectives without representing how the scene reprojects
under the motion of the agent. \textbf{(3) A policy deployed without the predictor.} The dynamics objectives
supervise the shared backbone during training only. Predicted states never enter action generation,
so the predictor is removed at deployment. The backbone and a
flow-matching action head then serve point-goal, image-goal, and goal-free navigation within a
unified model. Fig.~\ref{fig:difference} contrasts \skytopia with explicit and latent world models
on where the prediction is used.

Extensive experiments demonstrate that \skytopia achieves substantial improvements in monocular drone
navigation. In simulation, \skytopia surpasses seven competitive baselines on out-of-distribution (OOD) scenes
under all three navigation modes. It attains 57.8\%, 66.0\%, and 49.0\% success rate under the
point-goal, image-goal, and goal-free settings. This exceeds the strongest baseline by at least 15
points, on a 2B backbone and without scene-specific adaptation. Discarding the predictor after training removes 59.4\% of the inference
cost. We further validate \skytopia through real-world flight on a
physical drone in environments unseen during training, where the three modes attain 55\%, 56.7\%,
and 41.7\% success rate.

\section{Related Work}
\label{sec:related}

\paragraph{Learning-based navigation.} A large body of work replaces mapping and planning with a policy that maps observations directly to
actions. Early methods reached goals given as coordinates or as images \citep{shah2021recon,shah2022viking}, and later work scaled
across robots and datasets \citep{shah2023gnm,shah2023vint}. Since several trajectories are often equally valid, recent policies replace the regression head with a diffusion model
\citep{sridhar2024nomad,cai2025navdp,zeng2025navidiffusor,ren2025does}. On aerial platforms, learned
policies already operate at high speed in cluttered scenes
\citep{loquercio2021wild,kaufmann2023champion,zhang2025learning}. These aerial controllers typically
rely on depth sensing or on external localization, neither of which a monocular camera provides.
Across both ground and aerial settings, the only supervision available to such direct-mapping
policies is the demonstrated action. They fit the behaviour while leaving the dynamics that
produce it unmodelled, which makes them fragile under distribution shift.

\paragraph{Vision-language models for navigation.} A second line decodes actions directly from a
vision-language backbone that consumes the observation history as video
\citep{zhang2024navid,xu2024mobilityvla,hirose2025omnivla,zhang2026embodied}, and this formulation
has recently been extended to aerial agents \citep{zhang2025grounded,wang2025towards,gao2026openfly}.
Such models inherit strong semantic priors, and we adopt a vision-language backbone for the same
reason. However, instruction following is orthogonal to our setting, in which a goal is a
coordinate, an image, or absent. Moreover, these policies ground their representations in
appearance rather than in the geometry that flight requires, and their action generation remains a
direct mapping from observations to actions.

\paragraph{World models for navigation.} World models supervise how observations evolve under
actions, and navigation systems built on them fall into two categories. The first predicts future observations explicitly, synthesising with video generation the frames an agent would perceive
and passing them to a policy or to an inverse-dynamics module
\citep{bar2025nwm,luo2025grounding,chen2026imaginav,huang2026navdreamer,zhang2026sparse}. This
synthesis is computationally expensive, and the generated frames are often coherent in appearance
yet respond only weakly to the commanded action. The second predicts within a compact latent
space and plans over the imagined rollout
\citep{chahe2026policy,zhang2026rae,dong2026language,liu2026airdreamer,zhang2026hierarchical}, which
is substantially less costly but still executes the predictor at every control step. Within this second category, the strongest results to date are obtained by models that predict the future and
the action jointly in a single network \citep{yao2025navmorph,chen2025astranav,zhao2026worldvln,yang2026wam,zhou2026uninav,zhang2026futurenav}, which eliminates the separate rollout stage. Such systems
nevertheless target ground agents, and their action generation still depends on the predicted future
at deployment. Two systems approach our setting more closely: WorldVLN \mbox{\citep{zhao2026worldvln}} addresses aerial agents, and FutureNav
\citep{zhang2026futurenav} confines the dynamics objectives to training. The former, however,
executes a video backbone in closed loop and the latter regresses a pooled descriptor under a small
discrete action set. Neither is suited to a drone observing an open, weakly textured scene
through a single camera. \skytopia is designed for this setting: it acquires geometry from
prediction during training, and the deployed policy retains no predictor.

\section{Method}
\label{sec:method}

\subsection{Problem Formulation}
\label{sec:formulation}

We study offline policy learning for visual navigation.
The dataset $\mathcal{D}=\{(o_{0:T},s_{0:T},a_{0:T},g)\}$ contains trajectories of monocular
observations $o_t\in\mathbb{R}^{C\times H\times W}$, proprioceptive states
$s_t\in\mathbb{R}^{d_s}$, and body-frame velocity commands $a_t\in\mathbb{R}^{d_a}$.
A goal specification $g=(g^{\mathrm{p}},o^{\mathrm{g}})$ is supplied once per episode.
The metric goal vector $g^{\mathrm{p}}$ and goal image $o^{\mathrm{g}}$ are optional.
Their availability defines point-goal, image-goal, and goal-free navigation.
The policy produces an action chunk $A_t=(a_t,\dots,a_{t+N-1})$ over horizon $N$
and learns a latent representation of environment dynamics without reconstructing pixels.

\subsection{Overview}
\label{sec:overview}

\begin{figure}[t]
\centering
\includegraphics[width=0.95\textwidth]{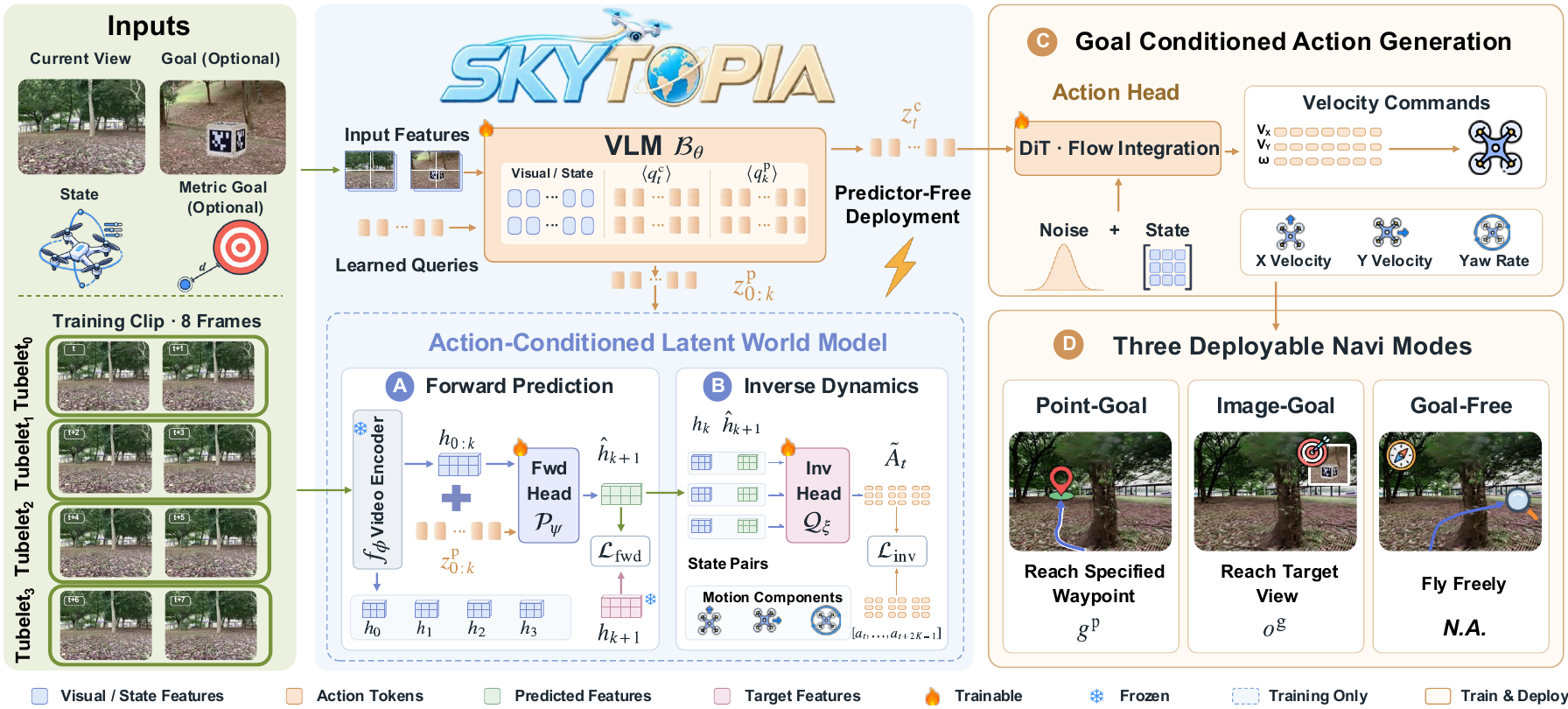}
\caption{\textbf{The \skytopia framework.} The backbone produces separate tokens for prediction
and control. During training, the forward head (A) predicts future latents and the inverse head (B)
recovers the executed motion. At deployment, only the backbone and flow-matching action head (C)
remain to support all three navigation modes (D).}
\label{fig:arch}
\end{figure}

Fig.~\ref{fig:arch} presents the \skytopia architecture. A shared vision--language backbone
encodes the current observation, proprioceptive state, and available goal inputs. Two groups of
learnable queries produce separate action tokens in a single backbone pass.
\textbf{Action tokens for prediction} encode motion over temporal windows and support two
training objectives. \textbf{Forward prediction} (A) estimates future latent states conditioned
on preceding states and these tokens. \textbf{Inverse dynamics} (B) recovers executed commands
from the predicted transitions. A frozen video encoder provides latent supervision from training
clips. \textbf{Action tokens for action generation} condition the \textbf{flow-matching action
head} (C) to generate body-frame velocity commands. The available goal inputs define
\textbf{point-goal, image-goal, and goal-free navigation} (D). Deployment retains only the
backbone and action head.

\subsection{Action-Conditioned Latent World Model}
\label{sec:wm}

\paragraph{Latent world state.} We supervise dynamics prediction in latent space without pixel
reconstruction. A pretrained video encoder $f_\phi$ is frozen to provide fixed targets:
\begin{equation}
h_k=\mathrm{sg}\!\left[f_\phi\!\left(o_{\mathcal{T}_k}\right)\right],
\label{eq:enc}
\end{equation}
where $\mathcal{T}_k$ denotes the $k$-th temporal window of the observation stream,
$\mathrm{sg}[\cdot]$ is the stop-gradient operator, and $h_k$ is the latent state of that window.
Latent states have a lower temporal resolution than control commands. A clip spanning $N$
commands yields $K+1$ states $h_{0:K}$ and $K<N$ transitions. The forward and inverse objectives
are defined over these transitions.

\paragraph{Action tokens.} Each temporal window $k$ is assigned a group of learnable queries
$\langle q^{\mathrm{p}}_k\rangle$. The backbone maps these queries to action tokens that encode
the intended motion:
\begin{equation}
z^{\mathrm{p}}_k=\mathcal{B}_\theta\!\left(\langle q^{\mathrm{p}}_k\rangle \,\middle|\, o_t,\,g,\,s_t\right),
\label{eq:ztok}
\end{equation}
where $\mathcal{B}_\theta$ denotes the backbone. Sec.~\ref{sec:goal} describes how the backbone
encodes $g$. The predictor uses the resulting tokens regardless of the goal modality.

\paragraph{Latent forward dynamics.} The forward head predicts the next latent state from
preceding states and action tokens:
\begin{equation}
\hat{h}_{k+1}=\mathcal{P}_\psi\!\left(h_{0:k},\,z^{\mathrm{p}}_{0:k}\right),
\label{eq:fwdpred}
\end{equation}
where $\mathcal{P}_\psi$ denotes the forward head and $\hat{h}_{k+1}$ is its prediction.
Attention is unrestricted within each window and causal across windows. The forward head
therefore cannot attend to future latent states. The action tokens $z^{\mathrm{p}}_{0:k}$ encode
the intended motion needed to predict viewpoint changes during flight. Preceding states alone
do not specify this motion.

We model the next latent state with a fixed-scale Laplace distribution centred at
$\hat{h}_{k+1}$. Maximizing its likelihood is equivalent to minimizing the $\ell_1$ prediction loss:
\begin{equation}
\mathcal{L}_{\mathrm{fwd}}=\sum_{k=0}^{K-1}\big\lVert \hat{h}_{k+1}-h_{k+1}\big\rVert_1 .
\label{eq:fwd}
\end{equation}
Future observations provide target states $h_{k+1}$ for dynamics supervision and are excluded
from the policy inputs in Eq.~\eqref{eq:ztok}. Freezing $f_\phi$ also prevents the target encoder
from collapsing to a constant representation that trivially minimizes the prediction loss.

\paragraph{Latent inverse dynamics.} Forward prediction alone provides weak supervision for
$z^{\mathrm{p}}_{0:k}$. Temporal correlations between latent states can reduce the prediction loss
without requiring action information. We therefore train an inverse head to recover executed
commands from predicted transitions:
\begin{equation}
\begin{aligned}
\tilde{A}_t&=\left[\mathcal{Q}_\xi\!\left(h_k,\,\hat{h}_{k+1}\right)\right]_{k=0}^{K-1},\\
\mathcal{L}_{\mathrm{inv}}&=\frac{1}{2Kd_a}\big\lVert \tilde{A}_t-[a_t,\ldots,a_{t+2K-1}]\big\rVert_1 ,
\end{aligned}
\label{eq:inv} \end{equation}
where $\mathcal{Q}_\xi$ operates on spatially pooled state pairs with shared weights. Brackets
denote concatenation. With tubelet size 2 and $K=3$, the inverse head recovers the first six
commands of the $N=7$ chunk. Only the policy objective supervises the final command $a_{t+6}$.
We randomly drop $h_{0:K-1}$ during training to limit reliance on preceding states and encourage
the use of $\hat{h}_{1:K}$. These predicted states propagate inverse-loss gradients through the
forward head to the backbone's action tokens. The two objectives thus jointly supervise latent
transitions and their associated commands. Derivations are provided in
Appendix~\ref{sec:appendix-derivations}; the dropout analysis is in
Appendix~\ref{sec:appendix-analysis}.

\subsection{Goal-Conditioned Action Generation}
\label{sec:goal}

\paragraph{Constructing the three navigation modes.} A metric goal is appended to the proprioceptive
state, whereas a goal image is presented to the backbone as an additional view. Formally,
\begin{equation}
z^{\mathrm{c}}_t=\mathcal{B}_\theta\!\left(\langle q^{\mathrm{c}}_t\rangle \,\middle|\, \mathcal{V}_t,\;\tilde{s}_t\right),
\qquad
\mathcal{V}_t=\{o_t\}\cup\{o^{\mathrm{g}}\!:m^{\mathrm{g}}\!=\!1\},
\qquad
\tilde{s}_t=\left[\,s_t,\;m^{\mathrm{p}}g^{\mathrm{p}}\,\right],
\label{eq:goal}
\end{equation}
where $\langle q^{\mathrm{c}}_t\rangle$ denotes the learnable queries and $z^{\mathrm{c}}_t$ denotes
the resulting action tokens. $\mathcal{V}_t$ contains the input views. The augmented state
$\tilde{s}_t$ combines $s_t$ with the masked metric goal. Binary indicators $m^{\mathrm{p}}$ and
$m^{\mathrm{g}}$ control the availability of each goal input. Both groups of action tokens are
computed in a single backbone pass. A metric goal defines point-goal navigation; a goal image
defines image-goal navigation. The goal-free mode omits both inputs.

The goal image retains its spatial features rather than being pooled into a single descriptor.
This preserves the spatial information needed to match regions between the current and goal
views. Appendix~\ref{sec:appendix-extra} evaluates this choice. During training, we sample
$(m^{\mathrm{p}},m^{\mathrm{g}})$ jointly from a categorical distribution over all four input
combinations. This exposes the policy to different goal specifications and discourages reliance
on a single goal modality.

\paragraph{Conditional flow-matching action head.} Obstacle avoidance can admit multiple valid
trajectories. Direct regression can average these alternatives into an unsuitable trajectory.
We use conditional flow matching to model a distribution over continuous action chunks and
evaluate this choice in Appendix~\ref{sec:appendix-extra}. Let $A_t$ denote the ground-truth chunk of $N$
commands. Let $\hat{A}_t$ denote the chunk generated by the learned flow. Following
\citet{intelligence2025pi_}, we define the interpolation
\begin{equation}
A^{\tau}_t=(1-\tau)\,\epsilon+\tau A_t , \qquad \tau=\frac{s-u}{s}, \qquad
u\sim\mathrm{Beta}(\alpha,\beta), \label{eq:interp} \end{equation} where
$\epsilon\sim\mathcal{N}(0,I)$ is Gaussian noise and $\tau$ interpolates between noise at $\tau=0$
and the demonstrated chunk at $\tau=1$. Beta sampling places greater weight on interpolation
times near the noise endpoint to emphasize the early stages of integration. The action head
parameterizes a vector field $v_\omega(A^{\tau}_t,\tau\mid z^{\mathrm{c}}_t,s_t)$ conditioned on
the action tokens and proprioceptive state. It is trained to match the velocity of the linear
interpolation:
\begin{equation}
\mathcal{L}_{\mathrm{act}}= \mathbb{E}_{A_t,\epsilon,\tau}\Big[\big\lVert
v_\omega\!\left(A^{\tau}_t,\tau\mid z^{\mathrm{c}}_t,s_t\right)-\left(A_t-\epsilon\right)\big\rVert_2^2\Big],
\label{eq:act} \end{equation} where $v_\omega(\cdot)$ denotes the predicted velocity field,
$(A_t-\epsilon)$ is the target velocity induced by the linear interpolation, and
$\lVert\cdot\rVert_2$ represents the $\ell_2$ norm. At inference time, the learned vector field is
integrated from noise to data space to obtain $\hat{A}_t$, after which the resulting commands are
executed before the next backbone query. Hyperparameters are reported in
Appendix~\ref{sec:appendix-training}.

\subsection{Training Objective and Deployment}
\label{sec:objective}

In summary, the overall training objective is as follows: \begin{equation}
\mathcal{L}=\mathcal{L}_{\mathrm{act}} +\lambda_{\mathrm{fwd}}\,\mathcal{L}_{\mathrm{fwd}}
+\lambda_{\mathrm{inv}}\,\mathcal{L}_{\mathrm{inv}} , \label{eq:total} \end{equation} which combines
Eqs.~\eqref{eq:act},~\eqref{eq:fwd} and~\eqref{eq:inv}, where $\lambda_{\mathrm{fwd}}$ and
$\lambda_{\mathrm{inv}}$ are tunable hyperparameters. Unlike world models whose rollout conditions
the policy at deployment, \skytopia never feeds $\hat{h}_{k+1}$ or $\tilde{A}_t$ into action
generation, and the dynamics objectives are not needed once training is complete. The deployed model
is therefore \begin{equation}
\hat{A}_t=v_\omega\!\left(\cdot\mid \mathcal{B}^{\,\mathrm{c}}_\theta\!\left(\mathcal{V}_t,\tilde{s}_t\right),\,s_t\right),
\label{eq:deploy}
\end{equation} where $\mathcal{B}^{\,\mathrm{c}}_\theta$ denotes the pathway of the backbone that produces
$z^{\mathrm{c}}_t$. The encoder $f_\phi$, the forward head $\mathcal{P}_\psi$, and the inverse head $\mathcal{Q}_\xi$
are all removed. The dynamics objectives therefore supervise the backbone without additional
inference cost. Architecture and training settings are detailed in
Appendices~\ref{sec:appendix-arch} and~\ref{sec:appendix-training}. Scene construction and
data collection are described in Appendix~\ref{sec:appendix-platform}.

\section{Experiments}
\label{sec:experiments}

We evaluate \skytopia on \textbf{unseen environments} under \textbf{three goal specifications}
and assess its \textbf{transfer to real-world flight}. Ablations and representation probes examine
the contribution of the dynamics objectives. We also compare \textbf{deployment efficiency}
against world-model baselines. All comparisons use matched observation and action spaces.

\subsection{Experimental Setup}
\label{sec:setup}

\paragraph{Benchmark.} All experiments run on the \skytopia platform, which comprises 94 scenes reconstructed from real
captures and simulated with rigid-body physics and scene-level collision,
split into 18 indoor spaces, 48 urban outdoor sites, and 28 vegetation scenes. We reserve 5 scenes
for OOD evaluation. Evaluation episodes use start--goal pairs sampled from mutually reachable
positions in free space. Each configuration is evaluated over 200 episodes per goal specification
under each of 3 random seeds. Scene construction and data collection are detailed in
Appendix~\ref{sec:appendix-platform}; evaluation procedures are in Appendix~\ref{sec:appendix-eval}.

\paragraph{Baselines.} We compare against the three categories in Sec.~\ref{sec:related}.
Policies that map observations directly to actions include behaviour cloning (BC)
\citep{codevilla2018end}, the action-chunking transformer ACT \citep{zhao2023learning},
NoMaD \citep{sridhar2024nomad}, and ViNT \citep{shah2023vint}. OmniVLA
\citep{hirose2025omnivla} represents vision--language--action policies with multimodal goal
conditioning. The world-model baselines are NWM \citep{bar2025nwm}, which generates future
observations, and NavMorph \citep{yao2025navmorph}, which predicts future latent states.
All baselines are adapted to monocular aerial navigation and retrained on the same data.

\paragraph{Metrics.} \label{sec:metrics} Performance is measured with the metrics that are standard in visual navigation \citep{anderson2018vln}. These are
Navigation Error (NE, m), Success Rate (SR, \%), Oracle Success Rate (OS, \%), Success weighted by
Path Length (SPL), Collision Rate (CR, \%), Out-of-Bounds Rate (OB, \%), and Time to Success (TTS). An episode is successful when the drone reaches the goal and
stops there without collision. Under goal-free navigation a goal is still sampled, but it is never
shown to the policy and serves only to compute the metrics. Definitions are given in
Appendix~\ref{sec:appendix-eval}.

\subsection{Simulation Results}
\label{sec:sim}

\begin{figure}[t]
\centering
\includegraphics[width=\textwidth]{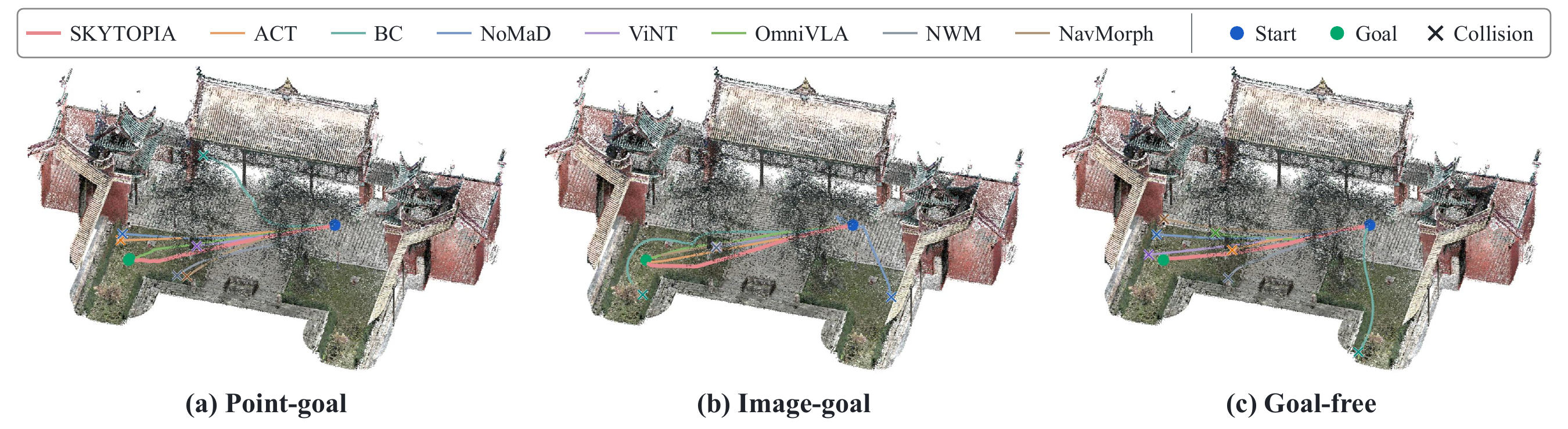} \caption{\textbf{Trajectory comparison on an OOD scene.} Executed trajectories of \skytopia
and of every baseline, drawn over the reconstruction and shown once per goal specification. Start
and goal are marked, and a cross marks the point where a run ends in collision.}
\label{fig:traj}
\end{figure}

\paragraph{Trajectory comparison.} Fig.~\ref{fig:traj} compares representative trajectories on
an unseen scene. \skytopia follows smooth paths and reaches the target in all three navigation
modes. Baseline trajectories exhibit collisions, unsuccessful goal approach, or detours with
abrupt changes in direction.

\paragraph{Main results.} Table~\ref{tab:main} shows that \skytopia achieves the highest SR,
OS, and SPL and the lowest NE and TTS under all three goal specifications. It consistently
outperforms OmniVLA, the strongest baseline, in both goal reaching and path efficiency while
substantially reducing collisions. The gains extend to goal-free navigation, where the policy
receives no target specification. Together, these results indicate improved obstacle avoidance
and efficient traversal in addition to more accurate goal approach.

We attribute these gains to \textbf{learning the relationship between motion and visual change}.
Forward prediction supervises how the scene evolves under the action tokens. Inverse dynamics
requires the predicted transitions to retain information about the executed commands. Their
combination encourages representations of scene geometry and camera motion that are useful for
obstacle avoidance across goal specifications. This interpretation is supported by the objective
ablations in Sec.~\ref{sec:ablation} and the geometry probes in Sec.~\ref{sec:analysis}.
Spatial goal features preserve local correspondences for goal alignment. The flow-matching head
models alternative manoeuvres without averaging their commands. The corresponding ablations
are reported in Appendix~\ref{sec:appendix-extra}.

\begin{table}[t]
\centering
\caption{\textbf{Closed-loop navigation on OOD scenes.} Results use 200 episodes per goal
specification per seed over 3 seeds. Best values are bold.}
\label{tab:main}
\vspace{2pt}
\scriptsize
\setlength{\tabcolsep}{2.2pt}
\resizebox{\textwidth}{!}{%
\begin{tabular}{l ccccccc ccccccc ccccccc}
\toprule
\multirow{2}{*}{Method} & \multicolumn{7}{c}{Point-Goal} & \multicolumn{7}{c}{Image-Goal} & \multicolumn{7}{c}{Goal-Free} \\
\cmidrule(lr){2-8}\cmidrule(lr){9-15}\cmidrule(lr){16-22}
& NE$\downarrow$ & OS$\uparrow$ & SR$\uparrow$ & SPL$\uparrow$ & CR$\downarrow$ & OB$\downarrow$ & TTS$\downarrow$
& NE$\downarrow$ & OS$\uparrow$ & SR$\uparrow$ & SPL$\uparrow$ & CR$\downarrow$ & OB$\downarrow$ & TTS$\downarrow$
& NE$\downarrow$ & OS$\uparrow$ & SR$\uparrow$ & SPL$\uparrow$ & CR$\downarrow$ & OB$\downarrow$ & TTS$\downarrow$ \\
\midrule
ACT & 2.99 & 24.2 & 23.3 & 0.17 & 73.2 & 3.5 & 145 & 2.68 & 35.2 & 34.3 & 0.25 & 62.8 & 2.8 & 136 & 3.17 & 17.2 & 14.3 & 0.09 & 82.5 & \textbf{3.2} & 152 \\
BC & 3.19 & 18.0 & 16.8 & 0.12 & 67.0 & 16.2 & 236 & 3.89 & 10.5 & 9.8 & 0.06 & 63.5 & 26.7 & 305 & 5.09 & 8.5 & 7.8 & 0.05 & \textbf{14.3} & 77.8 & 324 \\
NoMaD & 2.84 & 26.8 & 23.2 & 0.18 & 74.7 & 2.2 & 229 & 2.57 & 37.2 & 35.7 & 0.24 & 59.2 & 5.2 & 214 & 3.09 & 18.2 & 16.2 & 0.10 & 76.2 & 7.7 & 235 \\
ViNT & 3.03 & 22.2 & 19.8 & 0.14 & 74.2 & 6.0 & 235 & 2.94 & 30.2 & 27.3 & 0.20 & 60.2 & 12.5 & 224 & 3.23 & 16.3 & 15.2 & 0.09 & 77.3 & 7.5 & 246 \\
OmniVLA & 2.17 & 46.3 & 42.5 & 0.30 & 53.3 & 4.2 & 142 & 1.96 & 51.3 & 46.8 & 0.32 & 48.3 & 4.8 & 137 & 2.61 & 38.8 & 32.7 & 0.20 & 59.3 & 8.0 & 167 \\
NWM & 2.38 & 41.2 & 38.3 & 0.27 & 59.5 & 2.2 & 154 & 2.15 & 44.5 & 39.3 & 0.29 & 52.5 & 8.2 & 148 & 2.83 & 29.3 & 22.3 & 0.14 & 66.5 & 11.2 & 176 \\
NavMorph & 2.56 & 31.5 & 27.2 & 0.19 & 72.8 & \textbf{0.0} & 149 & 2.43 & 37.3 & 33.3 & 0.23 & 66.7 & \textbf{0.0} & 141 & 2.72 & 24.7 & 20.2 & 0.11 & 75.2 & 4.7 & 168 \\
\midrule
\ours \skytopia & \textbf{1.49} & \textbf{63.5} & \textbf{57.8} & \textbf{0.49} & \textbf{26.2} & 16.0 & \textbf{134} & \textbf{1.32} & \textbf{67.3} & \textbf{66.0} & \textbf{0.54} & \textbf{29.3} & 4.7 & \textbf{128} & \textbf{1.67} & \textbf{54.2} & \textbf{49.0} & \textbf{0.34} & 28.7 & 22.3 & \textbf{139} \\
\bottomrule
\end{tabular}}
\end{table}

\subsection{Real-World Flight}
\label{sec:real}

\begin{figure}[t]
\centering
\includegraphics[width=\textwidth]{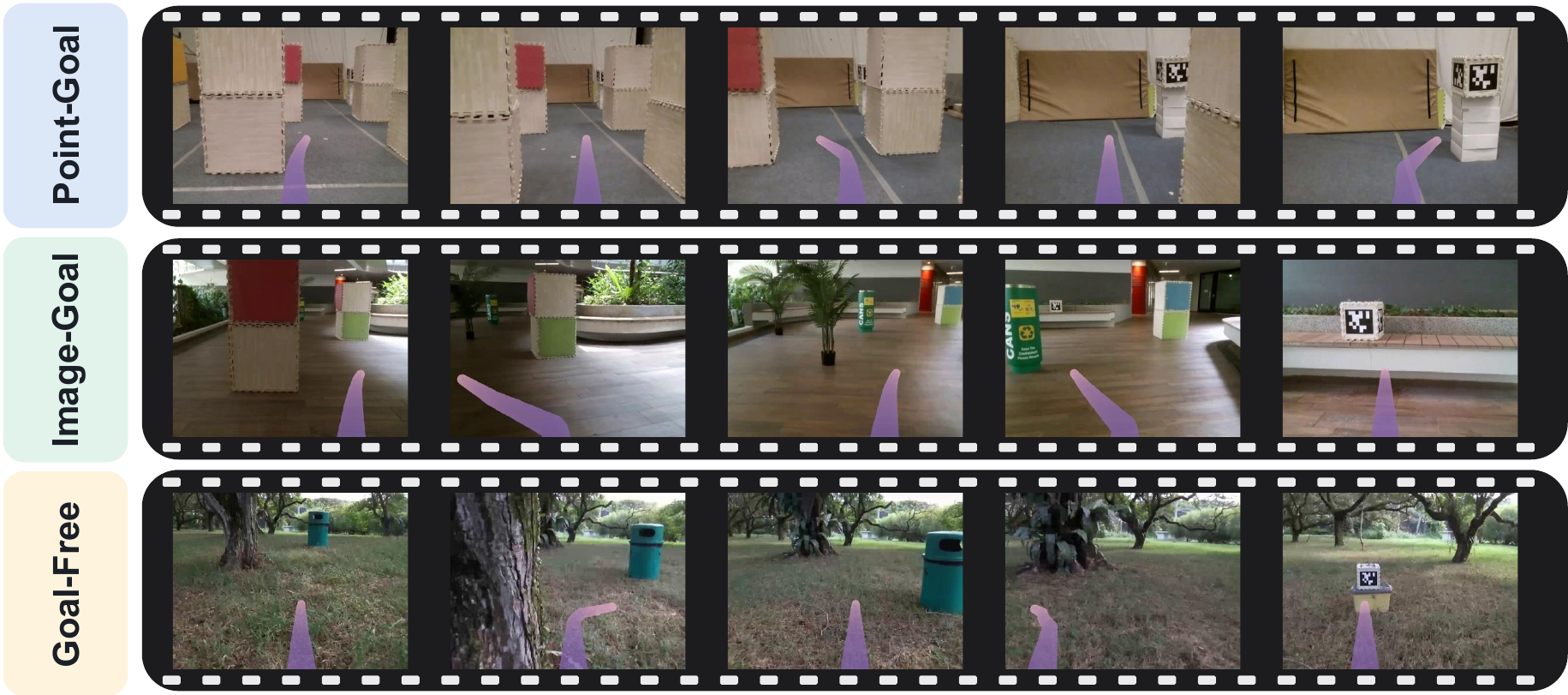}
\vspace{-8pt}
\caption{\textbf{Real-world flight.} Onboard first-person views from representative flights, one row per goal
specification and time running left to right. The overlay marks the direction the policy commands at
that moment.}
\label{fig:real}
\vspace{-6pt}
\end{figure}

We deploy the simulation-trained policy without fine-tuning in 3 unseen physical environments:
an indoor space, an open outdoor site, and woodland. Each setting uses 20 trials. Point-goal
navigation achieves 55\% SR indoors, where motion capture provides the metric goal. Image-goal
and goal-free navigation are evaluated at all 3 sites and achieve mean SRs of 56.7\% and 41.7\%,
respectively. Fig.~\ref{fig:real} shows representative flights with obstacle avoidance and
successful arrival under each goal specification. These results demonstrate transfer from
reconstructed scenes to physical flight across varied environments. Hardware and evaluation
details are provided in Appendix~\ref{sec:appendix-real}.

We attribute this transfer to \textbf{motion-conditioned representation learning} and
\textbf{training across diverse reconstructed scenes}. The forward and inverse objectives
encourage the backbone to encode spatial relationships relevant to flight instead of relying
solely on appearance--action associations. The training platform exposes this representation to
indoor, urban, and vegetation environments with varied geometry and appearance. Together, these
design choices provide a basis for generalization to unseen physical sites. The geometry probes
in Sec.~\ref{sec:analysis} and appearance perturbations in Appendix~\ref{sec:appendix-analysis}
provide supporting evidence for this interpretation.

\subsection{Ablation Studies}
\label{sec:ablation}

We isolate the contribution of the dynamics objectives using three variants:
w/o $\mathcal{L}_{\mathrm{fwd}}$ \& $\mathcal{L}_{\mathrm{inv}}$ uses only the action
objective; w/ $\mathcal{L}_{\mathrm{fwd}}$ adds forward prediction;
w/ $\mathcal{L}_{\mathrm{inv}}$ adds inverse dynamics. The latter retains the forward
head to produce the transitions used by the inverse objective. The full model uses both losses.
Table~\ref{tab:ablation} shows that removing both objectives reduces success and path efficiency
and increases collisions across all three modes. Either objective improves on action-only
training, while their combination gives the strongest overall navigation performance.

We attribute this pattern to the \textbf{complementary supervision of visual transitions and
executed motion}. Forward prediction constrains the predicted state to match the observed future,
but temporal correlations can weaken its dependence on action tokens. Inverse dynamics encourages
those tokens to encode motion through command recovery, but does not independently require the
predicted state to match the future observation. Joint training imposes both constraints. The
resulting gains in success and path efficiency support learning transitions that are visually
grounded and informative for control. Appendix~\ref{sec:appendix-extra} examines the goal
representation and action head separately.

\begin{table}[t]
\centering
\caption{\textbf{Ablation of forward and inverse dynamics objectives across three navigation modes.}}
\label{tab:ablation}
\vspace{2pt}
\scriptsize
\setlength{\tabcolsep}{2.2pt}
\resizebox{\textwidth}{!}{%
\begin{tabular}{l ccccccc ccccccc ccccccc}
\toprule
\multirow{2}{*}{Variant} & \multicolumn{7}{c}{Point-Goal} & \multicolumn{7}{c}{Image-Goal} & \multicolumn{7}{c}{Goal-Free} \\
\cmidrule(lr){2-8}\cmidrule(lr){9-15}\cmidrule(lr){16-22}
& NE$\downarrow$ & OS$\uparrow$ & SR$\uparrow$ & SPL$\uparrow$ & CR$\downarrow$ & OB$\downarrow$ & TTS$\downarrow$
& NE$\downarrow$ & OS$\uparrow$ & SR$\uparrow$ & SPL$\uparrow$ & CR$\downarrow$ & OB$\downarrow$ & TTS$\downarrow$
& NE$\downarrow$ & OS$\uparrow$ & SR$\uparrow$ & SPL$\uparrow$ & CR$\downarrow$ & OB$\downarrow$ & TTS$\downarrow$ \\
\midrule
w/o $\mathcal{L}_{\mathrm{fwd}}$ \& $\mathcal{L}_{\mathrm{inv}}$ & 2.05 & 42.2 & 37.5 & 0.25 & 52.7 & \textbf{9.8} & 158 & 1.87 & 46.8 & 42.8 & 0.29 & 52.5 & 4.7 & 161 & 2.32 & 33.2 & 29.2 & 0.18 & 58.2 & \textbf{12.7} & 157 \\
w/o $\mathcal{L}_{\mathrm{inv}}$ & 1.62 & 57.2 & 51.3 & 0.33 & 35.3 & 13.3 & 141 & 1.44 & 59.8 & 56.7 & 0.39 & 36.5 & 6.8 & 145 & 1.85 & 49.8 & 43.3 & 0.27 & 38.2 & 18.5 & 140 \\
w/o $\mathcal{L}_{\mathrm{fwd}}$ & 1.83 & 49.3 & 42.7 & 0.29 & 45.3 & 12.0 & 149 & 1.61 & 54.7 & 51.5 & 0.35 & 45.2 & \textbf{3.3} & 152 & 2.06 & 39.8 & 35.8 & 0.22 & 46.5 & 17.7 & 148 \\
\midrule
\ours \skytopia & \textbf{1.49} & \textbf{63.5} & \textbf{57.8} & \textbf{0.49} & \textbf{26.2} & 16.0 & \textbf{134} & \textbf{1.32} & \textbf{67.3} & \textbf{66.0} & \textbf{0.54} & \textbf{29.3} & 4.7 & \textbf{128} & \textbf{1.67} & \textbf{54.2} & \textbf{49.0} & \textbf{0.34} & \textbf{28.7} & 22.3 & \textbf{139} \\
\bottomrule
\end{tabular}}
\end{table}

\subsection{Analysis}
\label{sec:analysis}

\begin{figure}[t]
\centering
\begin{minipage}[t]{0.4400\textwidth}
\centering
\includegraphics[width=\textwidth]{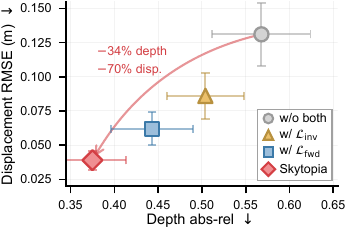}\\[-2pt]
{\small (a)}
\end{minipage}%
\begin{minipage}[t]{0.5600\textwidth}
\centering
\includegraphics[width=\textwidth]{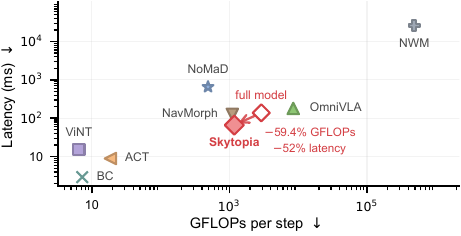}\\[-2pt]
{\small (b)}
\end{minipage}
\caption{\textbf{Geometric representation and inference cost.} (a) Depth and displacement probes
on frozen features. Markers and bars show means and standard deviations over 4 episode splits.
(b) Inference cost of every method. The arrow marks the effect of discarding the predictor
after training.}
\label{fig:analysis}
\end{figure}

\paragraph{Geometry in the learned representation.} We fit linear probes on frozen backbone
features to predict depth and camera displacement. Fig.~\ref{fig:analysis}(a) shows that joint
dynamics training yields lower errors on both tasks than the ablated variants. Geometry and
motion are therefore more accessible in the features without direct supervision of either
quantity during policy learning.

We attribute this result to the structure of the prediction task. Camera motion changes the
projected positions of scene elements according to their depth. Predicting these changes under
action conditioning encourages the representation to encode both scene structure and motion.
The inverse objective further requires the predicted transition to preserve command information.
The probe results therefore support the proposed connection between dynamics supervision and
the navigation improvements in Table~\ref{tab:ablation}.

\paragraph{Navigation performance and inference cost.} Fig.~\ref{fig:analysis}(b) reports
floating-point operations (FLOPs) and latency for every method. Discarding the predictor after
training removes 59.4\% of the FLOPs and 52\% of the latency, which moves \skytopia towards the
lower left. The deployed model then runs faster than NavMorph, OmniVLA, and NoMaD while reaching a
higher success rate than all of them across the three modes (Table~\ref{tab:main}). NWM requires
three orders of magnitude more computation and time per step. The methods that run faster than
\skytopia reach less than half its success rate.

This advantage follows from \textbf{restricting dynamics prediction to training}. The forward
and inverse objectives improve the backbone representation, while action generation uses a
separate token group and does not require predicted future states. Deployment therefore avoids
the computation of the dynamics branch. The comparison with NavMorph also shows that the
navigation gains do not require substantially more inference FLOPs.

\section{Conclusion}
\label{sec:conclusion}

We developed \skytopia, a monocular drone navigation framework that combines a 3DGS simulation
platform with action-conditioned latent world-model training. We use forward and inverse
dynamics objectives to learn a policy that supports three navigation modes without retaining
the predictor at deployment. Simulation experiments demonstrate improved goal reaching and
path efficiency over competitive baselines. We further deploy the same policy on a physical drone
without fine-tuning. Flights in unseen indoor, open outdoor, and woodland environments
demonstrate obstacle avoidance and successful arrival despite changes in appearance and flight
conditions. The policy supports metric goals indoors and both image-goal and goal-free navigation
across all three sites. These results establish the practical value of dynamics supervision for
sim-to-real monocular flight. However, the current policy lacks persistent spatial memory for long-horizon navigation, while its robustness to dynamic obstacles and temporally changing scenes remains unexplored. Future work will incorporate persistent memory for long-horizon reasoning and extend the framework to navigation in dynamic environments.

\subsection*{AI use statement}

We used generative AI tools to aid and polish writing. Their role was limited to grammar correction and to the rewording of text that the authors had already drafted. We did not
use generative AI tools for research ideation, method design, code implementation, experiment
execution, or the analysis of results. We have reviewed all AI-assisted text and confirmed that it
states our own claims and results. We take responsibility for the final content of this work,
including all text, claims, and artefacts.

\ifdefined\CaptureBranding
\subsubsection*{Acknowledgments}

We thank Autel US for providing the computational resources used in this work. We also thank
XGRIDS for providing the PortalCam handheld capture system with which the scenes of our platform
were reconstructed.
\fi

\bibliography{skytopia}
\bibliographystyle{iclr2027_conference}

\appendix
\section{Appendix}
\label{sec:appendix}

\subsection{Platform and Data Collection}
\label{sec:appendix-platform}

\begin{figure}[t]
\centering
\ifdefined\CaptureBranding
\includegraphics[width=\textwidth]{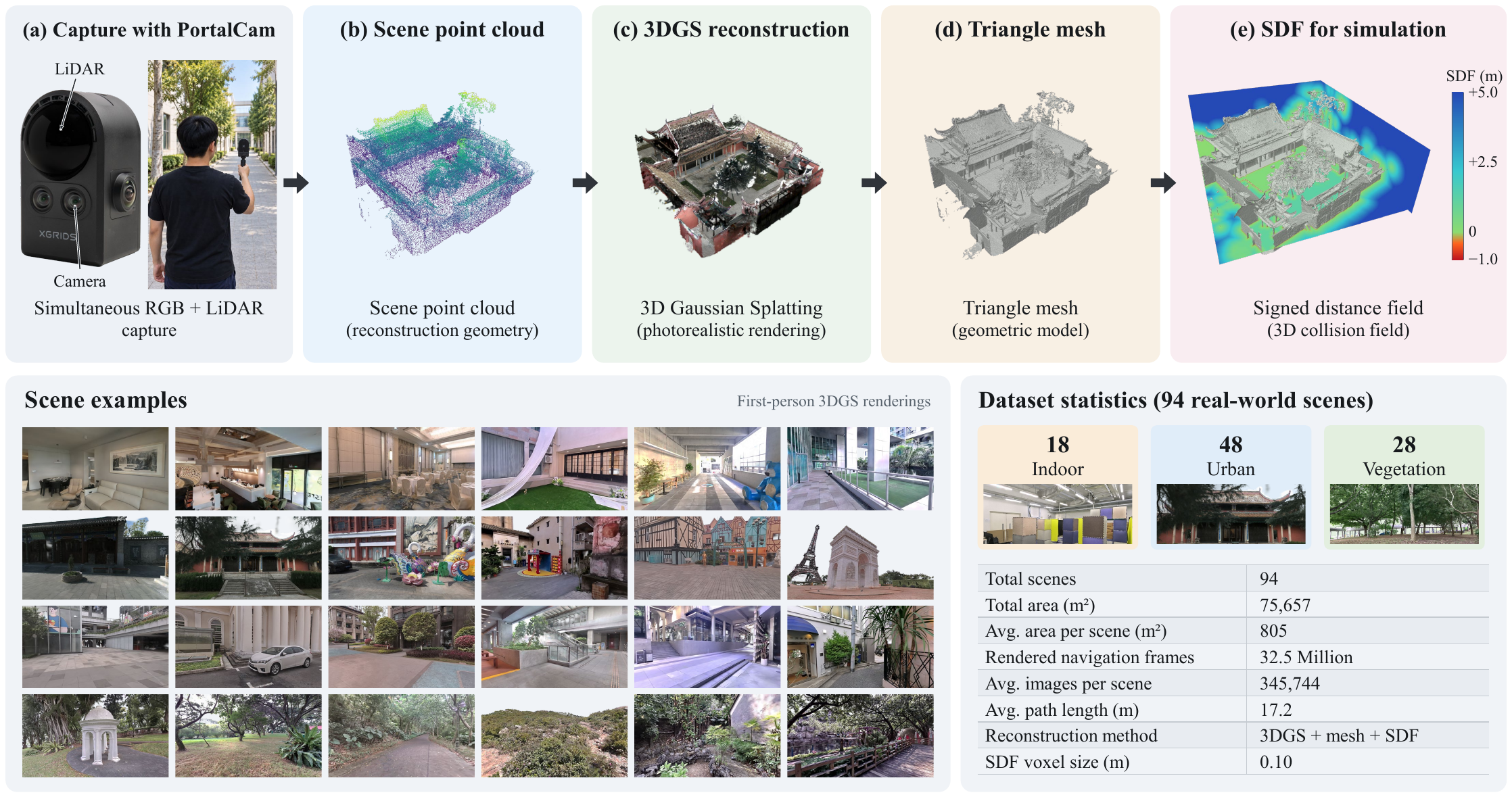}
\else
\includegraphics[width=\textwidth]{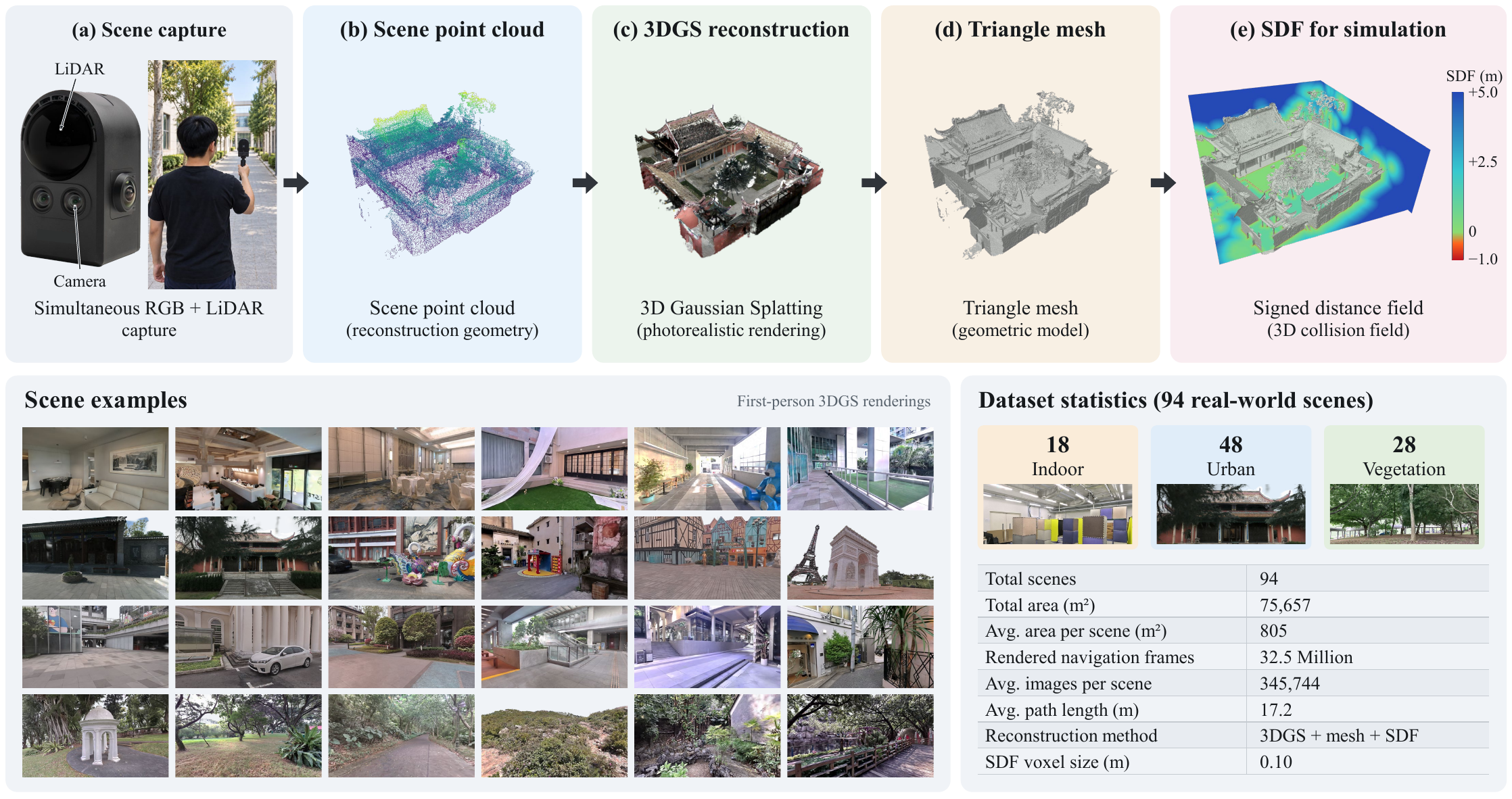}
\fi
\caption{\textbf{Scene reconstruction and the 3DGS simulation platform.} Top: synchronized
capture (a) provides the point cloud (b), 3DGS model (c), and mesh (d). The mesh supplies the
collision SDF (e). Bottom: 24 reconstructed scenes and dataset statistics.}
\label{fig:pipeline}
\end{figure}

\paragraph{Scene capture and reconstruction.} We capture each environment with a handheld
LiDAR--camera system that records synchronized point clouds and RGB images.
The LiDAR point cloud provides metric geometry. We optimize a 3DGS model against posed RGB
images to reproduce scene appearance. A triangle mesh extracted from the same reconstruction
provides the collision geometry. We convert this mesh into a signed distance field (SDF) with
$10\,$cm voxels for collision and clearance queries. The simulator uses 3DGS for rendering and
the SDF for physical interaction. Fig.~\ref{fig:pipeline} summarizes this pipeline.

The platform contains 94 scenes: 18 indoor spaces, 48 urban outdoor sites, and 28 vegetation
environments. These include rooms, corridors, streets, campus sites, gardens, and woodland.
The total navigable area is $75{,}657\,$m$^2$, with an average of $805\,$m$^2$ per scene.
Fig.~\ref{fig:pipeline} presents 24 example reconstructions and the platform statistics.

\paragraph{Data collection.} We import the reconstructed environments into Isaac Sim and
integrate rigid-body dynamics at $100\,\mathrm{Hz}$. Observations and commands are recorded at
$50\,\mathrm{Hz}$. The forward-facing camera renders $256\times256$ RGB images. Each action
$a_t=[v_x,\,v_y,\,\dot{\psi}]$ specifies body-frame velocity and yaw rate. The recorded
13-dimensional state contains linear velocity, angular velocity, the relative goal vector, and
the body orientation quaternion.

An expert planner generates demonstrations using the reconstructed scene geometry. It searches
for a shortest path on a mesh-derived occupancy grid and adjusts speed according to obstacle
clearance and path curvature. Yaw follows the path tangent with a rate limit of
$1.0\,$rad\,s$^{-1}$. The planner path also provides the reference length $\ell_i$ for SPL.
The collection contains 1000 trajectories per scene and 94{,}000 trajectories in total, yielding
$32.5$ million frames over $181$ hours of flight. The mean planner path length is $17.2\,$m,
and the mean episode length is $346$ control steps.

\subsection{Analysis of the Dynamics Objectives}
\label{sec:appendix-derivations}

\begin{figure}[t]
\centering
\includegraphics[width=\textwidth]{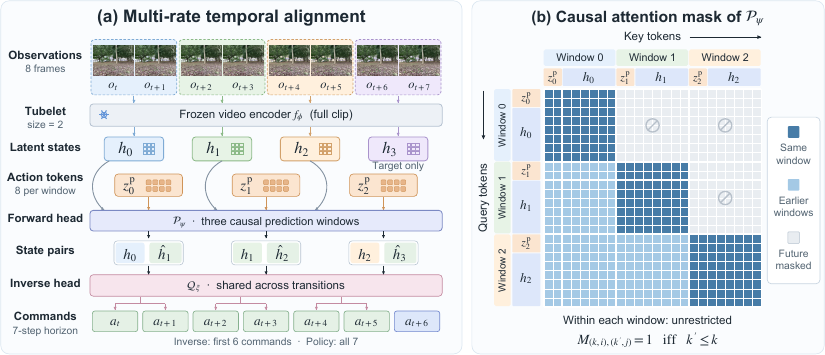}
\caption{\textbf{Temporal alignment and causal attention.} (a) The frozen encoder maps an
8-frame clip to 4 latent states using tubelet size 2. The forward head predicts 3 transitions
conditioned on 8 action tokens per transition. The shared inverse head recovers 2 commands from
each state pair, supervising the first 6 commands of the 7-command chunk. The action objective
supervises all 7 commands. (b) Tokens attend to the current and preceding windows; future
windows are masked.}
\label{fig:temporal}
\end{figure}

\paragraph{Temporal alignment.} Fig.~\ref{fig:temporal}(a) relates observations, latent states,
and commands. The frozen video encoder processes 8 frames with tubelet size 2 to produce 4
states. The backbone produces action tokens for the resulting $K=3$ transitions in one pass.
Each transition corresponds to 2 commands, giving inverse supervision for the first 6 commands
of the $N=7$ chunk. The action objective supervises all 7 commands, including the final command
without an inverse target.

\paragraph{Causal attention in the forward head.} Let $(k,i)$ denote token $i$ in window $k$.
The attention mask in $\mathcal{P}_\psi$ is
\begin{equation}
M_{(k,i),(k',j)}=
\begin{cases}
1, & k'\le k,\\[2pt]
0, & \text{otherwise.}
\end{cases}
\label{eq:mask}
\end{equation}
Tokens can attend to all positions in the current and preceding windows. The forward head cannot
attend to tokens from window $k+1$ or later when predicting $\hat{h}_{k+1}$.

\paragraph{Laplace likelihood and the $\ell_1$ objective.} We model the $d$ coordinates of each
latent state with conditionally independent Laplace distributions and a fixed scale $b>0$.
The distribution is centred at the forward prediction $\hat{h}_{k+1}$ from
Eq.~\eqref{eq:fwdpred}:
\begin{equation}
p_\psi\!\left(h_{k+1}\,\middle|\,h_{0:k},z^{\mathrm{p}}_{0:k}\right)
=\prod_{n=1}^{d}\frac{1}{2b}\exp\!\left(-\frac{\big|h^{(n)}_{k+1}-\hat{h}^{(n)}_{k+1}\big|}{b}\right),
\label{eq:laplace}
\end{equation}
Here $n$ indexes latent coordinates. Taking the negative logarithm converts the product into
a sum of absolute prediction errors. Summing over the $K$ transitions gives
\begin{equation}
-\sum_{k=0}^{K-1}\log p_\psi\!\left(h_{k+1}\,\middle|\,h_{0:k},z^{\mathrm{p}}_{0:k}\right)
=\frac{1}{b}\,\mathcal{L}_{\mathrm{fwd}}+Kd\log 2b .
\label{eq:nll}
\end{equation}
The term $Kd\log 2b$ is constant with respect to $\psi$ and $\theta$. Maximizing this likelihood
is therefore equivalent to minimizing Eq.~\eqref{eq:fwd}, with $1/b$ absorbed into
$\lambda_{\mathrm{fwd}}$. This probabilistic formulation motivates the loss without introducing
an additional uncertainty-prediction head. Absolute error grows linearly with the residual,
whereas squared error assigns disproportionately large penalties to large residuals. The
$\ell_1$ objective therefore limits their influence, which motivates its use when occlusion
or newly visible regions make parts of the next state difficult to predict.

\paragraph{Fixed targets and representation collapse.} Jointly optimizing the target encoder
and predictor admits the constant solution
\begin{equation}
f_\phi(\cdot)\equiv c \;\wedge\; \mathcal{P}_\psi(\cdot)\equiv c
\;\Longrightarrow\; \mathcal{L}_{\mathrm{fwd}}=0
\quad\text{for any constant } c ,
\label{eq:collapse}
\end{equation}
which minimizes the prediction loss without encoding dynamics. Both the target and prediction
are identical for every observation, and the loss cannot distinguish different transitions.
In \skytopia, $\partial\mathcal{L}_{\mathrm{fwd}}/\partial\phi\equiv 0$ by
Eq.~\eqref{eq:enc}. Training updates the predictor while target representations remain fixed.

Consider any two distinct fixed targets $h_i\ne h_j$. For a constant prediction $c$, the
triangle inequality gives
\begin{equation}
\lVert c-h_i\rVert_1+\lVert c-h_j\rVert_1
\ge\lVert h_i-h_j\rVert_1>0.
\label{eq:constant-bound}
\end{equation}
The same prediction cannot match both targets. Its combined loss is hence strictly positive.
Freezing the encoder thus prevents the target and predictor from jointly collapsing to a
zero-loss constant solution. This argument does not guarantee that the predictor uses action
tokens: correlated preceding states may still support prediction without them. The inverse
objective addresses this separate limitation by supervising command recovery from predicted
transitions.

\subsection{Architecture}
\label{sec:appendix-arch}

\paragraph{Modules.} We use Qwen3-VL-2B-Instruct \citep{bai2025qwen3} as the backbone
$\mathcal{B}_\theta$. The frozen target encoder $f_\phi$ is V-JEPA~2 ViT-L/16
\citep{assran2025v} with 300M parameters and $256\times256$ input resolution. It encodes
8-frame clips with tubelet size 2. The forward and inverse heads contain 160M and 2.1M
parameters, respectively. The action head is a 160M-parameter diffusion transformer (DiT) with 16 layers and hidden
width 768. Table~\ref{tab:arch} lists the complete configuration. Only the backbone and action
head are retained at deployment.

\paragraph{Action tokens.} We add two groups of special query tokens to the tokenizer and insert
them at fixed input positions. Prediction uses 8 tokens per transition for $z^{\mathrm{p}}_k$;
action generation uses 32 tokens per backbone pass for $z^{\mathrm{c}}_t$. The backbone's output
hidden states at these positions form the action tokens. The input prompt is
\emph{``Infer the temporal dynamics from frames \{actions\} and produce the corresponding policy
actions \{e\_actions\}. The last image is the goal view you must reach.''}
The placeholders contain the prediction and control queries, respectively. The final sentence
is included only for image-goal navigation to identify the goal view.

\begin{table}[t]
\centering
\caption{\textbf{Model architecture.} The final column indicates inference usage.}
\label{tab:arch}
\small
\setlength{\tabcolsep}{6pt}
\begin{tabular}{lll}
\toprule
Module & Configuration & Inference \\
\midrule
Backbone $\mathcal{B}_\theta$ & Qwen3-VL-2B-Instruct, 2B & yes \\
Action tokens $z^{\mathrm{p}}_k$ & 8 per transition, 24 per backbone query & no \\
Action tokens $z^{\mathrm{c}}_t$ & 32 per backbone query & yes \\
Observation resolution & $224\times224$ to $\mathcal{B}_\theta$, $256\times256$ to $f_\phi$ & yes \\
Video encoder $f_\phi$ & V-JEPA~2 ViT-L/16, 300M & no \\
Observation window & 8 frames, tubelet 2, 4 world states, $K=3$ & no \\
Forward head $\mathcal{P}_\psi$ & 12 layers, 8 heads, rotary position embedding, 160M & no \\
Inverse head $\mathcal{Q}_\xi$ & 2-layer multi-layer perceptron, 2.1M & no \\
Action head $v_\omega$ & DiT, 16 layers, width 768, 160M & yes \\
Action chunk & $N=7$, $d_a=3$ & yes \\
Proprioceptive state & $d_s=13$ & yes \\
Flow-matching steps & 4 & yes \\
\bottomrule
\end{tabular}
\end{table}

\subsection{Training Configuration}
\label{sec:appendix-training}

\paragraph{Optimization.} Table~\ref{tab:training} lists the optimizer, learning rates, and
loss weights. We assign separate learning rates to the backbone, action head, and dynamics heads.
Actions are normalized to $[-1,1]$ using the training-set extrema of each dimension. The same
statistics convert predicted actions back to physical units at deployment. Training uses 8 H200
GPUs for approximately 56 hours.

\paragraph{Goal sampling.} We sample the indicators $(m^{\mathrm{p}},m^{\mathrm{g}})$ jointly
for each training example. The combinations $(0,0)$, $(1,0)$, $(0,1)$, and $(1,1)$ have
probabilities $0.20$, $0.30$, $0.40$, and $0.10$, respectively. When $m^{\mathrm{g}}=1$, the
goal view is sampled $\Delta$ seconds after the current frame with
$\Delta\sim\mathcal{U}[0.5,4.0]$. Both views receive photometric jitter of strength $0.3$.

\begin{table}[t]
\centering
\caption{\textbf{Training configuration.}}
\label{tab:training}
\small
\setlength{\tabcolsep}{6pt}
\begin{tabular}{ll}
\toprule
Setting & Value \\
\midrule
Optimizer & AdamW, $\beta=(0.9,0.95)$, $\epsilon=10^{-8}$, weight decay $10^{-8}$ \\
Learning rate & $\mathcal{B}_\theta$: $1\times10^{-5}$, \; $v_\omega$: $1\times10^{-4}$, \; $\mathcal{P}_\psi,\mathcal{Q}_\xi$: $3\times10^{-5}$ \\
Schedule & cosine with minimum $10^{-6}$, 5000 warm-up steps \\
Training steps & 50000 \\
Batch size & 256 \\
Gradient clipping & 1.0 \\
$\lambda_{\mathrm{fwd}}$, $\lambda_{\mathrm{inv}}$ & 0.1, 0.1 \\
Dropout of $h_{0:K-1}$ in Eq.~\eqref{eq:inv} & 0.3 \\
Interpolation time $\tau$ & $(s-u)/s$, $u\sim\mathrm{Beta}(1.5,1.0)$, $s=0.999$, 1000 buckets \\
Flow-matching noise samples per example & 8 \\
\bottomrule
\end{tabular}
\end{table}

\subsection{Evaluation Protocol}
\label{sec:appendix-eval}

\paragraph{Termination criteria.} Evaluation uses closed-loop rollouts in held-out scenes.
Each episode terminates with success, collision, an out-of-bounds event, or a timeout after
3000 control steps. Success requires stopping within
$\varepsilon_{\mathrm{s}}=0.5\,\mathrm{m}$ of the goal without collision. The same distance
threshold is used in real-world evaluation. Collision is detected through the scene SDF;
an out-of-bounds event occurs when the drone leaves the reconstruction. Oracle success uses a
closest-approach threshold of $\varepsilon_{\mathrm{o}}=1.0\,\mathrm{m}$.

\paragraph{Metrics.} Let $M$ denote the number of episodes and $\mathbb{I}_i$ indicate success
in episode $i$. Let $d_i$ and $\ell_i$ denote the executed and planner path lengths,
respectively. NE measures terminal distance to the goal in metres. SR measures the fraction of
successful episodes, while OS measures the fraction that approach within
$\varepsilon_{\mathrm{o}}$ at any point. CR and OB measure collision and out-of-bounds
frequencies. These four rates are reported as percentages. TTS is the mean number of control
steps among successful episodes. Path efficiency is measured by
$\mathrm{SPL}=\frac{1}{M}\sum_{i=1}^{M}\mathbb{I}_i\,\ell_i/\max(d_i,\ell_i)$.

\paragraph{Image-goal specification.} The planner supplies a sequence of subgoals along its
reference path. The goal view $o^{\mathrm{g}}$ is updated as the drone advances, providing a
local visual target for routes where the final destination is initially outside the camera view.

\subsection{Baseline Adaptation}
\label{sec:appendix-baselines}

We adapt all baselines to aerial navigation and retrain them on the \skytopia dataset.
Each method receives a monocular RGB stream at its required resolution and is evaluated using
the protocol in Appendix~\ref{sec:appendix-eval}. All executed actions are body-frame velocity
commands. For methods that predict waypoints, we use the demonstration controller to convert
waypoints into velocity commands. This keeps waypoint tracking consistent across the relevant
baselines. Table~\ref{tab:baseline-details} summarizes the adaptations and computational costs.

\paragraph{Cost measurement.} We report floating-point operations (FLOPs) per policy query,
using the larger count from point-goal and image-goal inputs. One multiply-add counts as 2 FLOPs.
Parameter counts include frozen modules and inference auxiliaries.

\begin{table}[t]
\centering
\caption{\textbf{Aerial adaptations and computational cost of the baselines.}}
\label{tab:baseline-details}
\small
\setlength{\tabcolsep}{4pt}
\begin{tabular}{l p{0.42\linewidth} r r}
\toprule
Baseline & Adaptation & Params (M) & FLOPs (G) \\
\midrule
ACT~\mbox{\cite{zhao2023learning}} &
Retain conditional variational autoencoder (CVAE) action chunking;
add goal-conditioned image and state inputs and a 50-step drone velocity head.
& 83.89 & 18.66 \\
\midrule
BC~\mbox{\cite{codevilla2018end}} &
ResNet-18 features with state and goal fusion;
a multilayer perceptron (MLP) predicts velocity commands.
& 11.80 & 7.26 \\
\midrule
NoMaD~\mbox{\cite{sridhar2024nomad}} &
Retain action diffusion with image, state-history, and goal inputs;
predict body-frame velocity.
& 304.61 & 488.91 \\
\midrule
ViNT~\mbox{\cite{shah2023vint}} &
Retain visual encoding and distance and waypoint prediction;
add metric-goal and goal-free modes with aerial waypoint tracking.
& 31.47 & 6.48 \\
\midrule
OmniVLA~\mbox{\cite{hirose2025omnivla}} &
Retain the vision--language backbone and goal interface;
use low-rank adaptation (LoRA) and an aerial waypoint head.
& 7769.51 & 8471.54 \\
\midrule
NWM~\mbox{\cite{bar2025nwm}} &
Retain the world model and variational autoencoder (VAE).
Rank predictions with Learned Perceptual Image Patch Similarity (LPIPS).
Adapt aerial controls; use imagined metric-goal views and black goal images
for goal-free evaluation.
& 1098.08 & 485917.46 \\
\midrule
NavMorph~\mbox{\cite{yao2025navmorph}} &
Retain image encoding, spatiotemporal fusion, and Contextual Evolution
Memory (CEM). Build spatial features from image patches and add goal
conditioning and an aerial waypoint head.
& 317.77 & 1103.09 \\

\bottomrule
\end{tabular}
\end{table}

\subsection{Real-World Deployment Setup}
\label{sec:appendix-real}

\begin{wrapfigure}{r}{0.30\textwidth}
\centering
\includegraphics[width=0.92\linewidth]{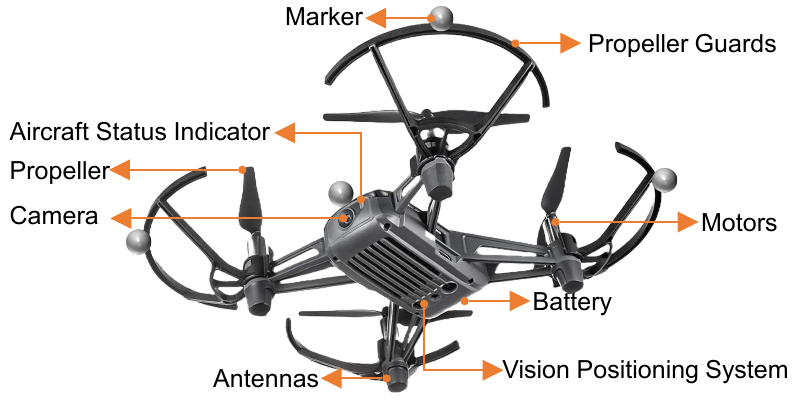}
\caption{The drone platform used in the real-world flights.}
\label{fig:platform}
\vspace{-0.5\baselineskip}
\end{wrapfigure}
\paragraph{Platform.} We use the DJI Tello shown in Fig.~\ref{fig:platform}. Its forward-facing
RGB camera streams 720p video at 30 FPS with a field of view of $82.6^\circ$. Indoor experiments
use an OptiTrack system with $120\,\mathrm{Hz}$ pose updates to compute the metric goal
$g^{\mathrm{p}}$. Outdoor experiments use the onboard inertial measurement unit (IMU) for body orientation and do not
receive motion-capture measurements.

\paragraph{Inference pipeline.} A WebSocket connection transmits the video stream to a ground
computer with an RTX 3090. We resize each frame to $224\times224$ and provide it to the policy
with the proprioceptive state. Inference runs at approximately $7\,\mathrm{Hz}$.
Real-time chunking \citep{black2026real} accounts for inference delay during command execution.
The commanded speed is limited to $1\,\mathrm{m\,s^{-1}}$.

\paragraph{Environments.} The indoor site is a $6\,\mathrm{m}\times6\,\mathrm{m}$ room with
box-shaped obstacles and potted plants. The open outdoor site spans approximately
$6\,\mathrm{m}\times20\,\mathrm{m}$ and contains scattered structures. The woodland site
covers approximately $30\,\mathrm{m}\times30\,\mathrm{m}$. Irregular trunks and low foliage
create narrow passages and partial occlusions. These sites evaluate transfer across differences
in appearance, obstacle layout, and visibility.

\paragraph{Goal specification and termination.} We manually specify start and goal positions
before each flight and retain the same routes across the modes supported at each site.
Point-goal navigation uses $g^{\mathrm{p}}$ and triggers landing on arrival. Image-goal
navigation uses visual subgoals recorded during a prior flight through the environment.
Landing is triggered when the cosine similarity between the current view and the final subgoal
exceeds a threshold. An operator judges arrival in goal-free navigation. The safety pilot ends
a flight after obstacle contact or departure from the test area. These events are recorded as
collision and out-of-bounds failures, respectively.

\subsection{Additional Ablations}
\label{sec:appendix-extra}

\begin{table}[t]
\centering
\caption{\textbf{Ablation of spatial goal features and the flow-matching action head on OOD scenes.}}
\label{tab:ablation-extra}
\vspace{2pt}
\scriptsize
\setlength{\tabcolsep}{2.2pt}
\resizebox{\textwidth}{!}{%
\begin{tabular}{l ccccccc ccccccc ccccccc}
\toprule
\multirow{2}{*}{Variant} & \multicolumn{7}{c}{Point-Goal} & \multicolumn{7}{c}{Image-Goal} & \multicolumn{7}{c}{Goal-Free} \\
\cmidrule(lr){2-8}\cmidrule(lr){9-15}\cmidrule(lr){16-22}
& NE$\downarrow$ & OS$\uparrow$ & SR$\uparrow$ & SPL$\uparrow$ & CR$\downarrow$ & OB$\downarrow$ & TTS$\downarrow$
& NE$\downarrow$ & OS$\uparrow$ & SR$\uparrow$ & SPL$\uparrow$ & CR$\downarrow$ & OB$\downarrow$ & TTS$\downarrow$
& NE$\downarrow$ & OS$\uparrow$ & SR$\uparrow$ & SPL$\uparrow$ & CR$\downarrow$ & OB$\downarrow$ & TTS$\downarrow$ \\
\midrule
w/ PG & 1.51 & 63.3 & 57.3 & 0.49 & 26.3 & 16.3 & 136 & 2.06 & 54.2 & 51.8 & 0.35 & 38.8 & 9.3 & 138 & 1.69 & 53.8 & 48.7 & 0.34 & 29.0 & 22.3 & 142 \\
w/ RH & 2.32 & 41.2 & 38.2 & 0.23 & 52.0 & \textbf{9.8} & 168 & 2.18 & 47.2 & 43.3 & 0.27 & 52.5 & \textbf{4.2} & 161 & 2.75 & 34.8 & 30.8 & 0.18 & 50.7 & \textbf{18.5} & 167 \\
\midrule
\ours \skytopia & \textbf{1.49} & \textbf{63.5} & \textbf{57.8} & \textbf{0.49} & \textbf{26.2} & 16.0 & \textbf{134} & \textbf{1.32} & \textbf{67.3} & \textbf{66.0} & \textbf{0.54} & \textbf{29.3} & 4.7 & \textbf{128} & \textbf{1.67} & \textbf{54.2} & \textbf{49.0} & \textbf{0.34} & \textbf{28.7} & 22.3 & \textbf{139} \\
\bottomrule
\end{tabular}}
\end{table}

\paragraph{Goal representation.} The w/ PG variant replaces spatial goal features with a pooled
descriptor. Table~\ref{tab:ablation-extra} shows a substantial reduction in image-goal success
and path efficiency, while point-goal and goal-free performance remain close to the full model.
The effect is therefore concentrated in the mode that uses the goal image.

We attribute this difference to the spatial information retained by the full representation.
Local correspondences between current and goal views help determine the motion needed for
alignment. Pooling removes the explicit spatial arrangement of goal features and makes these
correspondences less accessible. The results support preserving spatial features for visual
goal conditioning.

\paragraph{Action head.} The w/ RH variant replaces conditional flow matching with a regression
head. This change reduces success and path efficiency across all three modes
(Table~\ref{tab:ablation-extra}). The consistent decline indicates that the action-generation
objective contributes beyond the choice of goal input.

We attribute this advantage to modelling a distribution over valid action sequences. Obstacle
avoidance can admit distinct manoeuvres that a regression objective averages into an unsuitable
command sequence. Conditional flow matching represents these alternatives and can generate a
coherent action chunk. This helps explain the gains in success and path efficiency.

\subsection{Additional Analysis}
\label{sec:appendix-analysis}

\begin{figure}[t]
\centering
\includegraphics[width=\textwidth]{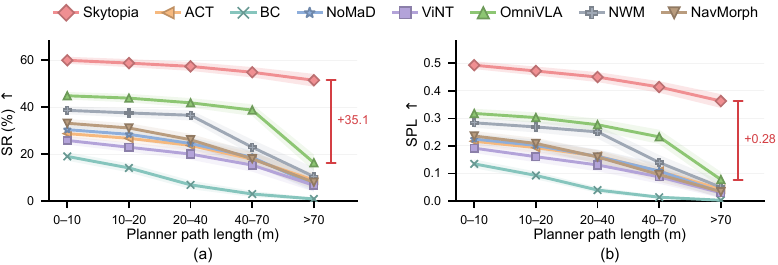}
\caption{\textbf{Navigation performance by route length.} Results are grouped by planner path
length and averaged over three goal modes. (a) SR. (b) SPL.
Bands show $\pm1$ standard deviation over 3 seeds. Brackets mark the margin over the strongest
baseline at the longest horizon.}
\label{fig:horizon}
\end{figure}

\paragraph{Navigation horizon.} We group episodes by planner path length to evaluate performance
over increasingly long routes. Fig.~\ref{fig:horizon} shows that \skytopia maintains higher
success and path efficiency across the evaluated horizons. Its advantage remains substantial
on the longest routes, where baseline performance declines more sharply.

We attribute this trend to the role of dynamics supervision in local navigation. Longer routes
require repeated obstacle avoidance and provide more opportunities for control errors to
accumulate. Representations that encode scene structure and motion can support more consistent
decisions at successive observations. The results are consistent with this interpretation.

\begin{figure}[t]
\centering
\includegraphics[width=0.96\textwidth]{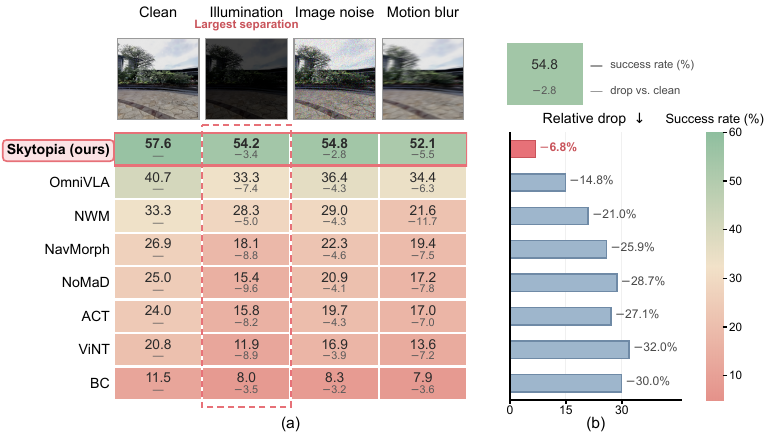}
\caption{\textbf{Robustness to visual perturbations.} Results are averaged over three goal
modes. (a) SR under each condition, with its decrease from the clean condition in points below.
Example observations appear above each column. (b) Mean decrease across the three perturbations,
expressed as a percentage of the clean SR.}
\label{fig:perturb}
\end{figure}

\paragraph{Visual perturbations.} We evaluate illumination changes, additive image noise, and
motion blur while preserving the underlying scene geometry. Each condition aggregates 1800
rollouts across the three goal modes. Fig.~\ref{fig:perturb} shows that \skytopia achieves the
highest SR under every perturbation and retains a larger fraction of its clean performance than
the baselines. Motion blur produces the largest decline for \skytopia, indicating that its
robustness remains sensitive to the quality of spatial image cues.

We attribute this robustness in part to dynamics supervision, which encourages features that
relate scene structure to camera motion. Such features can remain useful when appearance
changes without a corresponding change in geometry. The geometry probes in
Sec.~\ref{sec:analysis} support this interpretation. Blur weakens local boundaries and
correspondences needed for spatial reasoning, providing a plausible explanation for its larger
effect.

\Needspace*{9\baselineskip}
\begin{wraptable}{r}{0.44\textwidth}
\vspace{-\baselineskip}
\raggedleft
\caption{\textbf{Action recovery from frozen states.} Higher scores indicate better command
prediction. The aligned and unrelated pairs have equal input dimensions.}
\label{tab:probe}
\small
\setlength{\tabcolsep}{5pt}
\begin{tabular}{lc}
\toprule
Input & $1-\ell_1/\ell_1^{\mathrm{mean}}$ \\
\midrule
$(h_k,\,h_{k+1})$ & 0.791 \\
$h_k$ alone & 0.723 \\
$(h_k,\,h_{k'+1})$, unrelated $k'$ & 0.767 \\
\bottomrule
\end{tabular}
\end{wraptable}
\paragraph{Shortcut probe.} We fit action-recovery heads on 12{,}000 frozen windows and evaluate
them using $1-\ell_1/\ell_1^{\mathrm{mean}}$, where $\ell_1^{\mathrm{mean}}$ is the error of a
constant mean-command predictor. Table~\ref{tab:probe} compares an aligned state
pair, the preceding state alone, and a pair with an unrelated next state. The preceding state
alone predicts commands effectively. Replacing the next state with an unrelated one also
preserves much of the aligned pair's performance.

These results indicate that command recovery can exploit correlations in the preceding state
without relying strongly on the transition. To reduce this shortcut, we zero the preceding
state independently for each transition with probability 0.3 during inverse-head training.
Command recovery must then use the predicted next state, propagating gradients through the
forward head to the backbone.

\end{document}